\documentclass[sigconf,anonymous=False,review=False]{acmart}
\usepackage[utf8]{inputenc} % allow utf-8 input
\usepackage[T1]{fontenc}    % use 8-bit T1 fonts
\usepackage{hyperref}       % hyperlinks
\usepackage{url}            % simple URL typesetting
\usepackage{booktabs}       % professional-quality tables
\usepackage{amsfonts}       % blackboard math symbols
\usepackage{nicefrac}       % compact symbols for 1/2, etc.
\usepackage{microtype}      % microtypography
\usepackage{xcolor}         % colors
\usepackage{bbm}
\usepackage{mathtools}
\usepackage{natbib}
\usepackage{todonotes}
\usepackage{graphicx}
\usepackage{multirow}
\usepackage{algorithm}
\usepackage{algpseudocode}
\usepackage[skip=8pt]{caption}

\DeclareMathOperator*{\argmax}{arg\,max}
\DeclareMathOperator*{\argmin}{arg\,min}
\DeclareMathOperator*{\argsup}{arg\,sup}

\AtBeginDocument{%
  \providecommand\BibTeX{{%
    \normalfont B\kern-0.5em{\scshape i\kern-0.25em b}\kern-0.8em\TeX}}}

\setcopyright{acmcopyright}
\copyrightyear{2022}
\acmYear{2022}
\acmDOI{XXXXXXX.XXXXXXX}

\acmConference[Conference acronym 'XX]{Make sure to enter the correct
  conference title from your rights confirmation emai}{June 03--05,
  2018}{Woodstock, NY}
\acmPrice{15.00}
\acmISBN{978-1-4503-XXXX-X/18/06}

\newcommand{\repourl}{https://github.com/Mewtwo1996/ReLAX.git}

\makeatletter
  \if@ACM@anonymous
    \renewcommand{\repourl}{https://anonymous.4open.science/r/C9FB/README.md}
  \fi
\makeatother

\begin{document}

\title{{\sc ReLAX}: Reinforcement Learning Agent eXplainer for Arbitrary Predictive Models}

\author{Ziheng Chen}
\authornote{All the authors have contributed equally to this research.}
\affiliation{
  \institution{Stony Brook University, USA}
  %\country{USA}
  \country{}
}
\email{ziheng.chen@stonybrook.edu}

\author{Fabrizio Silvestri}
\affiliation{
  \institution{Sapienza University of Rome, Italy}
  \country{}
}
\email{fsilvestri@diag.uniroma1.it}

\author{Jia Wang}
\affiliation{
  \institution{The Xi'an Jiaotong-Liverpool University, China}
  \country{}
}
\email{jia.wang02@xjtlu.edu.cn}

\author{He Zhu}
\affiliation{
   \institution{Rutgers University, USA}
   \country{}
}
\email{hz375@cs.rutgers.edu}

\author{Hongshik Ahn}
\affiliation{
   \institution{Stony Brook University, USA}
   \country{}
}
\email{hongshik.ahn@stonybrook.edu}

\author{Gabriele Tolomei}
\affiliation{
  \institution{Sapienza University of Rome, Italy}
  \country{}
}
\email{tolomei@di.uniroma1.it}

\renewcommand{\shortauthors}{Chen, et al.}

\begin{abstract}
  Counterfactual examples (CFs) are one of the most popular methods for attaching post-hoc explanations to machine learning (ML) models. However, existing CF generation methods either exploit the internals of specific models or depend on each sample's neighborhood, thus they are hard to generalize for complex models and inefficient for large datasets.
  This work aims to overcome these limitations and introduces {\sc ReLAX}, a model-agnostic algorithm to generate optimal counterfactual explanations. Specifically, we formulate the problem of crafting CFs as a sequential decision-making task and then find the optimal CFs via deep reinforcement learning (DRL) with discrete-continuous hybrid action space.
  Extensive experiments conducted on several tabular datasets have shown that {\sc ReLAX} outperforms existing CF generation baselines, as it produces sparser counterfactuals, is more scalable to complex target models to explain, and generalizes to both classification and regression tasks.
  Finally, to demonstrate the usefulness of our method in a real-world use case, we leverage CFs generated by {\sc ReLAX} to suggest {\em actions} that a country should take to reduce the risk of mortality due to COVID-19. 
  Interestingly enough, the actions recommended by our method correspond to the strategies that many countries have actually implemented to counter the COVID-19 pandemic.
\end{abstract}






\maketitle

\newcommand{\tuple}[1]{(#1)}
\newcommand{\CFexp}{\textsc{CFExp}}
\newcommand{\R}{\mathbb{R}}
\newcommand{\E}{\mathbb{E}}
\newcommand{\X}{{\bf X}}
\newcommand{\dataset}{\mathcal{D}}
\newcommand{\features}{\mathcal{F}}
\newcommand{\inputs}{\mathcal{X}}
\newcommand{\labels}{\mathcal{Y}}
\newcommand{\params}{\boldsymbol{\theta}}
\newcommand{\weights}{\boldsymbol{\omega}}
\newcommand{\model}{m^*}
\newcommand{\loss}{\mathcal{L}}
\newcommand{\inst}{\boldsymbol{x}}
\newcommand{\cfinst}{\widetilde{\inst}}
\newcommand{\lpred}{\ell_{\text{pred}}}
\newcommand{\ldist}{\ell_{\text{dist}}}
\newcommand{\cfset}{\mathcal{C}_{\inst, h}}
\newcommand{\cfexp}{\boldsymbol{e}_{\inst}}
\newcommand{\states}{\mathcal{S}}
\newcommand{\actions}{\mathcal{A}}
\newcommand{\discount}{\gamma}
\newcommand{\trans}{\mathcal{T}}
\newcommand{\ptrans}{p_{\text{trans}}}
\newcommand{\mdp}{\mathcal{M} = \{\states, \actions, \trans, p_0, r, \discount\}}
\newcommand{\policy}{\pi:\states \mapsto P(\actions~|~\states)}
\newcommand{\return}{R_t = \sum_{i=0}^{\infty}\discount^{i}r(s_{t+i}, a_{t+i})}
\newcommand{\vpi}{V^{\pi}}
\newcommand{\qpi}{Q^{\pi}}
\newcommand{\svf}{\vpi(s) = \E(R_T~|~S_0=s;\pi)}
\newcommand{\asvf}{\qpi(s,a) = \E(R_T~|~S_0=s, A_0=a;\pi)}
\newcommand{\svfopt}{V^* = \text{sup}_{\pi}\vpi}
\newcommand{\asvfopt}{Q^* = \text{sup}_{\pi}\qpi}

\newcommand{\bellmaneqV}{V^*(s) = \max\limits_{a\in \actions} \mathop{\E}_{s'\sim \trans(s,a,\cdot)}[r(s,a,s') + \discount V^*(s')]}

\newcommand{\bellmaneqQ}{Q^*(s,a) = \mathop{\E}_{s'\sim \trans(s,a,\cdot)}[r(s,a,s') + \discount \max\limits_{a'\in \actions} Q^*(s', a')]}

\newcommand{\dqnloss}{\loss_t(\weights) = \{Q(s_t, a_t; \weights) - [r_t + \discount \text{max}_{a'\in \actions} Q(s', a'; \weights_t)]\}^2}

\newcommand{\policygrad}{\nabla_{\params}J(\pi_{\params}) = \E_{s,a} [\nabla_{\params} \log \pi_{\params}(a|s)Q^{\pi_{\params}}(s, a)]}

%%%%%%%%%%%%%%%%%%%%%%%%%%%%% INTRODUCTION %%%%%%%%%%%%%%%%%%%%%%%%%%%%%
\section{Introduction}
\label{sec:introduction}

% CONTEXT
Recent years have witnessed surprising advances in machine learning (ML), which in turn have led to the pervasive application of ML models across several domains. 
Unfortunately, though, many ML systems deployed in the wild are treated as ``black boxes'', whose complexity often hides the inner logic behind their output predictions. 
In fact, knowing why an ML model returns a certain output in response to a given input is pivotal for a variety of reasons, such as model debugging, aiding decision-making, or fulfilling legal requirements~\cite{gdpr}.

% IMPACT ON THE COMMUNITY
To properly achieve model transparency, a new initiative named {\em eXplainable AI} (XAI) has emerged~\cite{samek2019explainableAI}. 
A large body of work on XAI has flourished in recent years~\cite{guidotti2018lore}, and approaches to XAI can be broadly categorized into two classes~\cite{verma2020arxiv}: \emph{(i) native} and \emph{(ii) post-hoc}. The former leverages ML models that are inherently interpretable and transparent, such as linear/logistic regression, decision trees, association rules, etc. The latter aims at generating {\em ex post} explanations for predictions made by opaque or black-box models like random forests and (deep) neural networks.

% OUR FOCUS
In this work, we focus on specific post-hoc explanations called {\em counterfactual explanations}, which are used to interpret predictions of individual instances in the form: ``{\em If}  A {\em had been different,} B {\em would not have occurred}''~\cite{stepin2021survey}\cite{wachter2017hjlt}. 
They work by generating modified versions of input samples that result in alternative output responses from the predictive model, i.e., {\em counterfactual examples} (CFs).

% APPROACHES TO COUNTERFACTUAL EXAMPLES GENERATION
Typically, the problem of generating CFs is formulated as an optimization task, whose goal is to find the ``closest'' data point to a given instance, which crosses the decision boundary induced by a trained predictive model.\footnote{We voluntarily left this notion of ``closeness'' underspecified here; a more precise definition of it is provided in Section~\ref{sec:cf-problem}.}
Depending on the level of access to the underlying predictive model, different CF generation methods have been proposed.
More specifically, we can broadly distinguish between three categories of CF generators: \emph{(i) model-specific}~\cite{kentaro2020dace, karimi2020neurips, karimi2020aistats, russell2019fat, ustun2019fat}, \emph{(ii) gradient-aware}~\cite{wachter2017hjlt, pawelczyk2020www}, and \emph{(iii) model-agnostic}~\cite{dandl2020ppsn}.
% LIMITATIONS OF EXISTING CF GENERATORS
In short, existing CF generators require complete knowledge of the model's internals (e.g., random forests or neural networks), or they work only with differentiable models, or finally, they depend on each sample's neighborhood. Thus, they are hard to generalize for more complex models and inefficient for large datasets.

To overcome these limitations, we introduce a novel CF generation method called {\sc ReLAX}.
Specifically, we formulate the problem of crafting CFs as a sequential decision-making task. We then find the optimal CFs via deep reinforcement learning (DRL). The intuition behind {\sc ReLAX} is the following. 
The transformation of a given input instance into its optimal CF can be seen as the sequence of actions that an agent must take in order to get the maximum expected reward (i.e., the optimal policy). The total expected reward considers both the desired CF prediction goal (i.e., the CF and the original sample must result in different responses when they are input to the predictive model) and the distance of the generated CF from the original instance. At each time step, the agent has either achieved the desired CF transformation or it needs to: \emph{(i)} select a feature to modify and \emph{(ii)} set the magnitude of such change.
To generate meaningful CFs, the agent must restrict itself to operate on the set of {\em actionable} features only, as not every input feature can always be changed (e.g., the ``{\em age}'' of an individual cannot be modified).
Moreover, even if actionable, some features can only be changed toward one direction (e.g., a person can only increase their ``{\em educational level}''). 
Plus, the total number of tweaked features, as well as the magnitude of the change, must also be limited, as this would likely result in a more feasible CF.
We show that solving the constrained objective to find the optimal CF generator is equivalent to learning the optimal policy of a DRL agent operating in a discrete-continuous hybrid action space.

% OUR MAIN FINDINGS AND RESULTS
Our proposed method is a model-agnostic CF generator, as it can be applied to {\em any} black-box predictive model.
Indeed, in our framework the predictive model is just a ``placeholder'' resembling the environment which the DRL agent interacts with.

We validate our method on five datasets (four of them used for classification and one for regression), and compare it against several baselines using four standard quality metrics.
Experimental results show that {\sc ReLAX} outperforms all the competitors in every single metric.
To further demonstrate the impact of our CF generation method in practice, we show that CFs generated by {\sc ReLAX} can be leveraged to suggest {\em actions} that a country should take to reduce the risk of mortality due to the COVID-19 pandemic.

Overall, the main contributions of this work are as follows:

% SUMMARY OF THE MAIN CONTRIBUTIONS

\begin{itemize}
    % FIRST DRL APPROACH TO CF EXPLANATIONS
    \item {\sc ReLAX} is the first method for generating model-agnostic counterfactual examples based on deep reinforcement learning. It is scalable with respect to the number of features and instances. It can explain {\em any}  black-box model trained on tabular input data, regardless of its internal complexity and prediction task (classification or regression). 
    % ReLACE-Global vs. ReLACE-Local
    \item We implement two variants of our method: {\sc ReLAX-Global} and {\sc ReLAX-Local}, with the latter obtained via transfer learning from a pretrained version of the former.
    Both methods generate +60\% valid CFs that are about 40\% sparser than those produced by state of the art techniques yet take 42\% less CF generation time.
    % CURIOSITY
    \item To overcome the sparse reward problem due to the large state and action space, we integrate a hierarchical curiosity-driven exploration mechanism into our DRL agent.
    % PRACTICAL IMPACT
    \item We further assess the power of {\sc ReLAX} on a real-world use case. More specifically, we show that the actions suggested by {\sc ReLAX} to reduce the risk of mortality due to the COVID-19 pandemic correspond to the strategies that many countries have actually put in practice.
    % SOURCE CODE
    \item The source code implementation of {\sc ReLAX} and the datasets (including the one used for the COVID-19 risk of mortality task) are made publicly available.\footnote{\url{\repourl}}
\end{itemize}

% PAPER ORGANIZATION/STRUCTURE
The remainder of the paper is organized as follows. 
In Section~\ref{sec:related-work}, we review related work.
In Section~\ref{sec:cf-problem}, we formalize the problem of generating counterfactual explanations, while in Section~\ref{sec:framework} we present {\sc ReLAX}, our proposed method to solve this problem using deep reinforcement learning. 
We validate our approach and discuss the main findings of our work in Section~\ref{sec:experiments}.
In Section~\ref{sec:casestudy}, we further demonstrate the practical impact of {\sc ReLAX} on a real-world use case.
Finally, Section~\ref{sec:conclusion} concludes the paper.
%%%%%%%%%%%%%%%%%%%%%%%%%%%%%%%%%%%%%%%%%%%%%%%%%%%%%%%%%%%%%%%%%%%%%%%%
%%%%%%%%%%%%%%%%%%%%%%%%%%%%% RELATED WORK %%%%%%%%%%%%%%%%%%%%%%%%%%%%%
% RELATED WORK
\section{Related Work}
\label{sec:related-work}

\subsection{Counterfactual Explanations for Machine Learning Predictions}
A comprehensive review of the most relevant work in this area can be found in~\cite{verma2020arxiv}\cite{guidotti2022counterfactual}\cite{wang2021skyline}\cite{poyiadzi2020face}\cite{lucic2022aistats}.
A possible approach to counterfactual explanation is called {\sc Nearest-CT}~\cite{le2020grace}. Instead of generating a synthetic CF for a given input sample, this approach selects the nearest CF data point from the training set. More sophisticated counterfactual explanation methods can be broadly classified into {\em model-specific} and {\em model-agnostic}; as the names suggest, the former are tailored for a particular ML model (e.g., random forest), whereas the latter are able to generate explanations for any model.

\noindent{\bf \em Model-specific.}
One of the first counterfactual explanation method -- {\sc FeatTweak} -- is proposed by Tolomei et al.~\cite{tolomei2017kdd}, which is specifically designed for random forests, and exploits the internal structure of the learned trees to generate synthetic counterfactual instances. Another approach that is conceived for explaining tree ensembles is called FOCUS~\cite{lucic2019focusAAAI}. It frames the problem of finding counterfactual explanations as an optimization task and uses probabilistic model approximations in the optimization framework. 
Meanwhile, the rise of deep learning has given way to more complex and opaque neural networks (NNs).
In this regard, {\sc DeepFool}~\cite{moosavi2016deepfool} -- which was originally designed for crafting adversarial examples to undermine the robustness of NNs -- has proven effective also as a CF generation method.
However, CFs obtained from adversarial techniques often require changing almost all the features of the original instances, making them unrealistic to implement. Thus, Le et al.~\cite{le2020grace} propose GRACE: a novel technique that explains NN model's predictions using sparser CFs, which therefore are suitable also for high-dimensional datasets.

\noindent{\bf \em Model-agnostic.}
% LORE
Guidotti et al.~\cite{guidotti2018lore} introduce LORE, which first trains an interpretable surrogate model on a synthetic sample's neighborhood generated by a genetic algorithm. Then, LORE derives an explanation consisting of a decision rule and a set of counterfactual rules.
% MACE
More recently, Karimi et al.~\cite{karimi2020aistats} propose MACE, which frames the generation of model-agnostic CFs into solving a sequence of satisfiability problems, where both the distance function (objective) and predictive model (constraints) are represented as logic formulae.
% DiCE
In addition, Mothilal et al.~\cite{mothilal2020fat} present DiCE, a framework for generating and evaluating a diverse and feasible set of counterfactual explanations, which assumes knowledge of the model's gradients is available, thus not fully model-agnostic.

Our {\sc ReLAX} method is totally model-agnostic and aims to be as general and flexible as possible. 
Moreover, different from existing model-agnostic methods, {\sc ReLAX} is much more efficient to generate optimal CFs.
Indeed, {\sc ReLAX} requires to train a DRL agent that makes use only of the input/output nature of the target predictive model to explain, regardless of its internal complexity or its gradients (as opposed to DiCE).
Our method better scales to high-dimensional, large datasets than LORE: the genetic algorithm used to build each synthetic sample's neighborhood may be unfeasible for large feature spaces. 
Plus, LORE also requires to train a locally-interpretable decision tree that is tight to each generated neighborhood, and therefore may be prone to overfitting.
{\sc ReLAX} can also seamlessly handle more complex models than MACE (e.g., deeper NNs), which needs to construct a first-order logic characteristic formula from the predictive model and test for its satisfiability. 
This may be intractable when the formula (i.e., the model to explain) is too large.  
Finally, in contrast with {\sc ReLAX}, both LORE and MACE do not consider nor control over the sparsity of the generated CFs; moreover, LORE does not even take into account their actionability.

\subsection{Reinforcement Learning with Parameterized Action Space}
Many real-world reinforcement learning (RL) problems requires complex controls with discrete-continuous hybrid action space.
For example, in {\em Robot Soccer}~\cite{masson2016reinforcement}, the agent not only needs to choose whether to shoot or pass the ball (i.e., discrete actions) but also the associated angle and force (i.e., continuous parameters).
Unfortunately, most conventional RL algorithms cannot deal with such a heterogeneous action space directly.
The straightforward methods either discretize the continuous action space into a large discrete set~\cite{sherstov2005function},  or convert a discrete action into a continuous action method~\cite{hausknecht2016half}, but they significantly increase the problem complexity.
To overcome this issue, a few recent works propose to learn RL policies over the original hybrid action space directly.
Specifically, they consider a parameterized action space containing a set of discrete actions $A=\{a_1,a_2,\ldots,a_{|A|}\}$ and corresponding continuous action-parameter $v_k\in V_{k}\subseteq\R$. 
In this way, the action space can be represented as:
$\actions=\bigcup_{k}\{(a_k,v_k)~|~a_k\in A, v_k\in V_{k}\}.$
Masson et al.~\cite{masson2016reinforcement} propose a learning framework Q-PAMDP that alternatively learns the discrete action selection via $Q$-learning and employs policy search to get continuous action-parameters. Following the idea of Q-PAMDP, Khamassi et al.~\cite{khamassi2017active} treat two actions separately. The only difference is that they use policy gradient to optimize the continuous parameters. 
Both methods are on-policy and assume that continuous parameters are normally distributed. Also, Wei et al.~\cite{wei2018hierarchical} propose a hierarchical approach to deal with the parameterized action space, where the parameter policy is conditioned on the discrete policy.  
Although efficient, this method is found to be unstable due to its joint-learning nature. 
Recently, in order to avoid approximation as well as reduce complexity, Xiong et al.~\cite{xiong2018parametrized} introduce P-DQN, which seamlessly combines and integrate both DQN~\cite{mnih2013nipsw} and DDPG~\cite{lillicrap2016iclr}. Empirical study indicates that P-DQN is efficient and robust.
%%%%%%%%%%%%%%%%%%%%%%%%%%%%%%%%%%%%%%%%%%%%%%%%%%%%%%%%%%%%%%%%%%%%%%%%
%%%%%%%%%%%%%%%%%%%%%%%%%%%%% PROBLEM %%%%%%%%%%%%%%%%%%%%%%%%%%%%%
\section{Problem Formulation}
\label{sec:cf-problem}
% PRELIMINARIES
Let $\inputs \subseteq \R^n$ be an input feature space and $\labels$ an output label space. Without loss of generality, we consider both the $K$-ary {\em classification} setting, i.e., $\labels = \{0, \ldots, K-1\}$, and the {\em regression} setting, i.e., $\labels \subseteq \R$.
Suppose there exists a predictive model $h_{\weights}: \inputs \mapsto \labels$, parameterized by $\weights$, which accurately maps any input feature vector $\inst = (x_1,\ldots,x_n) \in \inputs$ to its label $h_{\weights}(\inst)=y\in \labels$.

% IDEA
The idea of counterfactual explanations is to reveal the rationale behind predictions made by $h_{\weights}$ on individual inputs $\inst$ by means of counterfactual examples (CFs). 
More specifically, for an instance $\inst$, a CF $\cfinst \neq \inst$ according to $h_{\weights}$ is found by perturbing a subset of the features of $\inst$, chosen from the set $\features \subseteq \{1,\ldots,n\}$.
% CF PREDICTION GOAL
The general goal of such modification is to transform $\inst$ into $\cfinst$ so to change the original prediction, i.e., $h_{\weights}(\cfinst) \neq h_{\weights}(\inst)$~\cite{wachter2017hjlt}.
In particular, this depends on whether $h_{\weights}$ is a classifier or a regressor.
In the former case, the objective would be to transform $\inst$ into $\cfinst$ so to change the original predicted class $c = h_{\weights}(\inst)$ into another $\widetilde{c} = h_{\weights}(\cfinst)$, such that $\widetilde{c} \neq c$.
Notice that $\widetilde{c}$ can be either specified upfront (i.e., \emph{targeted} CF) or it can be {\em any} $\widetilde{c} \neq c$ (i.e., {\em untargeted} CF). 
In the case $h_{\weights}$ is a regressor, instead, the goal is trickier: one possible approach to specifying the validity of a counterfactual example $\cfinst$ is to set a threshold $\delta \in \R\setminus \{0\}$ and let $|h_{\weights}(\cfinst) - h_{\weights}(\inst)| \geq \delta$. 
However, CFs found via such a thresholding are %known to be very 
sensitive to the choice of $\delta$~\cite{spooner2021arxiv}.

% OPTIMAL CF
Either way, as long as the CF classification or regression goal is met, several CFs can be generally found for a given input $\inst$.
This may lead to a situation where many CFs are in fact unrealistic or useless, as they are too far from the original instance.
Therefore, amongst all the possible CFs, we search for the {\em optimal} $\cfinst^*$ as the ``closest'' one to $\inst$.
Intuitively, this is to favor CFs that require the {\em minimal} perturbation of the original input.

More formally, let $g_{\params}:\inputs\mapsto \inputs$ be a counterfactual generator, parameterized by $\params$, that takes as input $\inst$ and produces as output a counterfactual example $\cfinst = g_{\params}(\inst)$.
For a given sample $\dataset$ of i.i.d. observations drawn from a probability distribution, i.e., $\dataset \sim p_{\text{data}}(\inst)$, we can measure the cost of generating CFs with $g_{\params}$ for all the instances in $\dataset$, using the following counterfactual loss function:
\begin{equation}
\label{eq:cf-loss}
\loss_{\text{CF}}(g_{\params};\dataset,h_{\weights}) = \frac{1}{|\dataset|}\sum_{\inst \in \dataset} \lpred(\inst, g_{\params}(\inst); h_{\weights}) + \lambda \ldist(\inst, g_{\params}(\inst)).
\end{equation}
The first component ($\lpred$) penalizes when the CF prediction goal is {\em not} satisfied. 
Let $\mathcal{C}\subseteq \inputs$ be the set of inputs which do {\em not} meet the CF prediction goal, e.g., in the case of classification, $\mathcal{C} = \{\inst\in \inputs~|~h_{\weights}(\inst) = h_{\weights}(g_{\params}(\inst))\}$.
Hence, we can compute $\lpred$ as follows:
\begin{equation}
\label{eq:lpred}
\lpred(\inst, g_{\params}(\inst); h_{\weights}) = \mathbbm{1}_{\mathcal{C}}(\inst),
\end{equation}
where $\mathbbm{1}_{\mathcal{C}}(\inst)$ is the well-known 0-1 indicator function, which evaluates to 1 iff $\inst \in \mathcal{C}$, 0 otherwise.

The second component $\ldist:\inputs\times \inputs\mapsto \R_{>0}$ is any arbitrary distance function that discourages $\cfinst$ to be too far away from $\inst$.
For example, $\ldist(\inst, g_{\params}(\inst)) = ||\inst - g_{\params}(\inst)||_p$, where $||\cdot||_p$ is the $L^p$-norm.
In this work, inspired by previous approaches, we set $\ldist$ equal to $L^1$-norm~\cite{guidotti2018lore}.

In addition, $\lambda$ serves as a scaling factor to trade off between $\ldist$ and $\lpred$.
Notice, though, that not every input feature can be lightheartedly modified to generate a valid CF, either because it is strictly impossible to do it (e.g., the ``{\em age}'' of an individual cannot be changed), and/or due to ethical concerns (e.g., the ``{\em race}'' or the ``{\em political tendency}'' of a person).
Therefore, we must restrict $\features$ to the set of {\em actionable} features only.
Plus, the total number of perturbed features must also be limited, i.e., $|\features| \leq m$ for some value $1\leq m\leq n$.

\begin{figure}[h]
\centering
\includegraphics[width=\columnwidth]{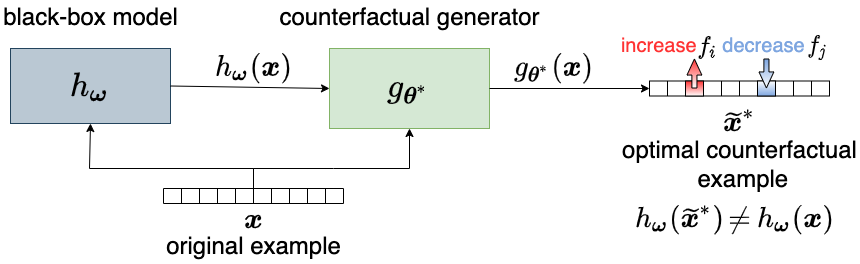}
\caption{Overview of a generic counterfactual explainer.}
\label{fig:cf-generic}
\end{figure}

Eventually, we can find the optimal CF generator $g^* = g_{\params^*}$ as the one whose parameters $\params^*$ minimize Equation~\ref{eq:cf-loss}, i.e., by solving the following constrained objective:
\begin{equation}
\label{eq:optimalcf}
\begin{aligned}
\params^*~=~& \argmin\limits_{\params}\Big\{\loss_{\text{CF}}(g_{\params};\dataset,h_{\weights})\Big\} \\
& \textrm{subject to:} \quad |\features| \leq m.
\end{aligned}
\end{equation}
This in turn allows us to generate the optimal CF $\cfinst^*$ for any $\inst$, as $\cfinst^* = g^*(\inst)$.
Finally, the resulting optimal counterfactual explanation can be simply computed as $\cfexp = \cfinst^* - \inst$~\cite{tolomei2021tkde}. 
The overview of a generic counterfactual explanation framework is shown in Figure~\ref{fig:cf-generic}.

%%%%%%%%%%%%%%%%%%%%%%%%%%%%%%%%%%%%%%%%%%%%%%%%%%%%%%%%%%%%%%%%%%%%%%%%
%%%%%%%%%%%%%%%%%%%%%%%%%%%%% ReLAX %%%%%%%%%%%%%%%%%%%%%%%%%%%%%
\section{Proposed Framework: {\sc R\lowercase{e}LAX}}
\label{sec:framework}

In this work, we propose to find the optimal CF generator for any arbitrary model -- i.e., to solve the constrained optimization problem defined in Equation~\ref{eq:optimalcf} -- via deep reinforcement learning (DRL).
We call our method \emph{\textbf{Re}inforcement \textbf{L}earning \textbf{A}gent e\textbf{X}plainer} ({\sc ReLAX}).

\subsection{Markov Decision Process Formulation}
We consider the problem of computing the optimal counterfactual example $\cfinst^*$ from $\inst \in \dataset$ -- i.e., the optimal CF generator $g^*$ defined in Equation~\ref{eq:optimalcf} -- as a sequential decision-making task.
More precisely, we refer to the standard reinforcement learning setting, where at each time step an agent: \emph{(i)} takes an action (i.e., selects a feature of the original sample $\inst$ to tweak {\em and} the magnitude of such change) and \emph{(ii)} receives an observation (i.e., the prediction output by $h_{\weights}$ on the input just modified according to the action taken before) along with a scalar reward from the environment.
The process continues until the agent eventually meets the specified CF prediction goal and the optimal CF $\cfinst^*$ is found.

We formulate this process as a standard Markov Decision Process (MDP) $\mdp$. In the following, we describe each component of this framework, separately.

% States
\subsubsection{States ($\states$)}
At each time step $t$, the agent's state is $S_t=s_t$, where $s_t = (\inst_t, \boldsymbol{f}_t)\in \states$ represents the current modified sample $(\inst_t)$ along with the set of features changed so far ($\boldsymbol{f}_t$). 
More specifically, $\boldsymbol{f}_t\in \{0,1\}^{|\features|}$ is an $|\features|$-dimensional binary indicator vector, where $\boldsymbol{f}_t[k] = 1$ iff the actionable feature $k$ has been modified in one of the actions taken by the agent {\em before} time $t$.
Initially, when $t=0$, $\inst_0 = \inst$ and $\boldsymbol{f}_0 = 0^{|\features|}$.

If the prediction goal is met, e.g., $h_{\weights}(\inst_t) \neq h_{\weights}(\inst)$, the agent reaches the end of the episode and the process terminates returning $\cfinst^* = \inst_t$ as the CF for $\inst$.
Otherwise, the agent must select an action $A_t=a_t$ to: {\em{(i)}} pick a feature to change amongst those which have not been modified yet and {\em{(ii)}} decide the magnitude of that change.

% Discrete-Continuous Action Space
\subsubsection{Discrete-Continuous Hybrid Actions ($\actions$)}
To mimic the two-step behavior discussed above, we consider a discrete-continuous hybrid action space with a two-tier hierarchical structure.

For an arbitrary step $t$, we maintain the set of feature identifiers that the agent is allowed to modify $\features_t$; initially, when $t=0$, $\features_0 = \features$ as the agent can pick {\em any} of the actionable features to change.
Then, at each time step $t > 0$, the agent first chooses a high level action $k_t$ from the discrete set $\features_t \subset \features = \features \setminus \bigcup_{j=0}^{t-1}k_j$.
This is to allow each feature to be selected at most in one action.
Upon choosing $k_t\in \features_t$, the agent must further select a low level parameter $v_{k_t} \in \R$, which specifies the magnitude of the change applied to feature $k_t$. It is worth noticing that we can confine the action's range and direction by directly putting a constraint on the low level parameter $v_{k_t}$.
Overall, $a_t = (k_t, v_{k_t})$, and we define our discrete-continuous hybrid action space as follows:
\[
\actions_t = \{(k_t, v_{k_t})~|~k_t\in \features_t, v_{k_t}\in \R\}.
\]
% Transition Function
\subsubsection{Transition Function ($\trans$)}
Let $a_t = (k_t, v_{k_t})\in \actions_t$ be the generic action that the agent can take at time $t$.
The action $a_t$ deterministically moves the agent from state $s_t$ to state $s_{t+1}$, by operating on $\inst_t$ and $\boldsymbol{f}_t$ as follows:
\begin{equation}
\label{eq:transcf}
\trans((\inst_t, \boldsymbol{f}_t), a_t, (\inst_{t+1}, \boldsymbol{f}_{t+1})) = \begin{cases}
1,\text{ if }\inst_{t} \overset{\mathclap{a_t}}{\leadsto} \inst_{t+1}  \wedge  \boldsymbol{f}_{t} \overset{\mathclap{a_t}}{\leadsto} \boldsymbol{f}_{t+1} \\
0, \text{ otherwise}.
\end{cases}
\end{equation}
The statements $\inst_{t} \overset{\mathclap{a_t}}{\leadsto} \inst_{t+1}$ and $\boldsymbol{f}_{t} \overset{\mathclap{a_t}}{\leadsto} \boldsymbol{f}_{t+1}$ are shorthand for $\inst_{t+1}[k_t] = \inst_t[k_t] + v_{k_t}$ and  $\boldsymbol{f}_{t+1}[k_t] = 1$, respectively.
This corresponds to increasing the value of feature $k_t$ by the magnitude $v_{k_t}$, and updating the binary indicator vector $\boldsymbol{f}_t$ accordingly.

% Reward
\subsubsection{Reward ($r$)}
The reward is expressed in terms of the objective function defined in Equation~\ref{eq:optimalcf} and has the following form:
\begin{equation}
\label{eq:rewardcf}
r(s_t, a_t) =
\begin{cases}
1-\lambda (\ldist^t-\ldist^{t-1}),\text{ if }h_{\weights}(\inst_t) \neq h_{\weights}(\inst)\\
-\lambda (\ldist^t-\ldist^{t-1}), \text{ otherwise},
\end{cases}
\end{equation}
where $\ldist^t = \ldist(\inst, \inst_t)$ and $\lambda \in \R_{>0}$ is a parameter that controls how much weight to put over the distance between the current modified instance ($\inst_t$) and the original sample ($\inst$).
In other words, the agent aims to reach a trade-off between achieving the CF prediction goal and keeping the distance of the counterfactual from the original input sample $\inst$ as lower as possible.

% Policy
\subsubsection{Policy ($\pi_{\params}$)}
We define a {\em parameterized} policy $\pi_{\params}$ to maximize the expected reward in the MDP problem. 
Our ultimate goal, though, is to find an optimal policy $\pi^* = \pi_{\params^*}$ that solves Equation~\ref{eq:optimalcf}.
It is worth noticing that finding the optimal policy $\pi^*$ that maximizes the expected return in this environment is equivalent to minimizing Equation~\ref{eq:cf-loss}, and thereby finding the optimal CF generator $g^*$.
The equivalence between these two formulations is shown below:
\begin{equation*}
\begin{aligned}
\params^{*} &=\argmin_{\params}\frac{1}{|\dataset|}\sum_{\inst \in \dataset} \lpred(\inst, g_{\params}(\inst); h_{\weights}) + \lambda \ldist(\inst, g_{\params}(\inst))\\
 &=\argmax_{\params}\frac{1}{|\dataset|}\sum_{\inst \in \dataset}-\lpred(\inst, g_{\params}(\inst);h_{\weights}) -\lambda \ldist(\inst, g_{\params}(\inst))\\
 &=\argmax_{\params}\frac{1}{|\dataset|}\sum_{\inst \in \dataset}\begin{cases}
1-\lambda (\ldist^t-\ldist^{t-1}),\text{ if }h_{\weights}(\inst_t) \neq h_{\weights}(\inst)\\
-\lambda (\ldist^t-\ldist^{t-1}), \text{ otherwise},
\end{cases} \\
 &=\argmax_{\params}\frac{1}{|\dataset|}\sum\limits_{\inst\in \dataset}\sum\limits_{t=1}^{T}r(s_t,\pi_{\params}(s_t)).%T(\inst)
\end{aligned}
\end{equation*}
Here, $T$ defines the maximum steps taken by the agent for each sample $\inst$ and is set to 50,000.

\subsection{Policy Optimization}
\label{subsec:policy-opt}
We use the P-DQN framework~\cite{xiong2018parametrized} to find the optimal policy $\pi^*$. 
At each time step, the agent takes a hybrid action $a_t\in \actions_t$ to perturb the currently modified $\inst_t$, obtained from the original input $\inst$.
\begin{figure*}[ht]
\centering
\includegraphics[width=.66\textwidth]{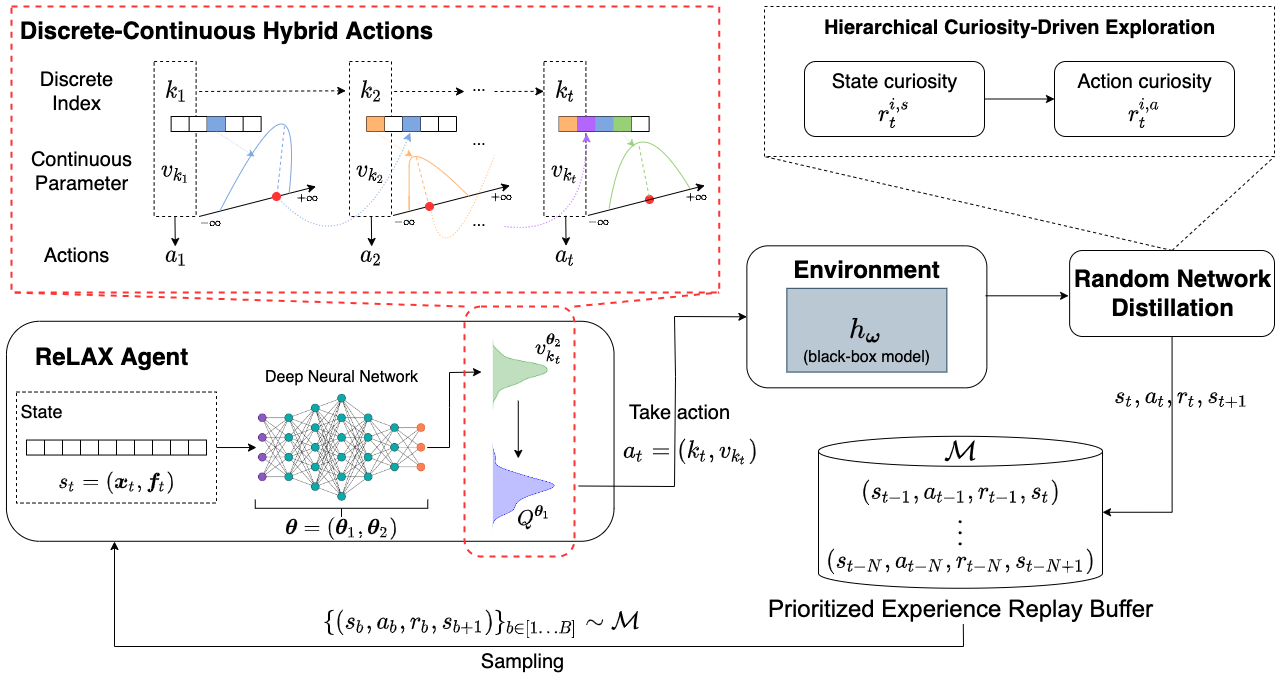}
\caption{Overview of our proposed {\sc ReLAX} framework.}
\label{fig:relace}
\end{figure*}
In $Q$-learning, one aims at finding the optimal $Q$-value function representing the expected discounted reward for taking action $a_t$ at a given state $s_t$.
Inside the $Q$-value function, the continuous parameter $v_{k_t}$ is associated with the discrete action $k_t$, which means $v_{k_t}$ is the optimal action given state $s_t$ and $k_t$: $v_{k_t}=\argsup_{v}Q(s_{t+1},k_t,v)$.
We therefore cast this as a function $v_{k_t}^{Q}(s_t)$. Thus, the Bellman equation can be written as:
\begin{equation*}
    Q(s_t,k_t,v_{k_t}) = \mathbb{E}_{r_t,s_{t+1}}(r_t+\gamma \max_{k_t}  Q(s_{t+1},k_t,v^{Q}_{k_t}(s_{t+1}))|s_t,a_t=(k_t,v_{k_t})).
\end{equation*}
As with DQN, a deep neural network (DNN) $Q^{\params_1}(s_t,k_t,v_{k_t})$ is used to approximate the $Q$-value function, and we fit $v_{k_t}^{Q}$ with another deterministic policy network $v_{k_t}^{\params_2}(s_t)$, where $\params_1$ and $\params_2$ are the parameters of the two DNNs. 
To find the optimal $\params_1^*, \params_2^*$, we minimize the loss functions below via stochastic gradient descent:
\[\mathcal{L}_Q(\params_1)=[Q^{\params_1}(s_{t},k_t,v_{k_t})-y_t]^2,~~~\mathcal{L}_{\pi}(\params_2)=-\sum\limits_{k_t\in\features}Q(s_t,k_t,v_{k_t}^{\params_2}(s_t)),\] where the $y_t$ is the n-step target~\cite{sutton1998rl}. Theoretically, as the error diminishes, the $Q$ network converges to the optimal $Q$-value function and the policy network outputs the optimal continuous action. However, this method is unstable in practice. This is because DQN only samples transition pairs uniformly from the replay buffer. Therefore, we adopt prioritized experience replay \cite{schaul2015prioritized} to effectively learn from pairs with high expected learning value. As a measurement for learning potential, prioritized experience replay samples transition pairs with probability $p_j$ based on their TD error:
$p_t \propto |R_j+\gamma Q_{\mbox{target}}(s_j,Q(s_j,a_j))-Q(s_{j-1},a_{j-1})|^{\beta}$,
 where $\beta$ is a parameter to determine how much prioritization is
used. 
During training, the transition pairs are stored in the replay buffer with their priority; a sum tree data structure is then used to efficiently update the priority and sample pairs from the replay buffer.

\subsection{Hierarchical Curiosity-Driven Exploration }
\label{subsec:Curiosity}
In the procedure of generating counterfactual examples, the sparse reward problem is inevitably met due to the large state and action space. Reward shaping is a typical solution that converts the sparse reward to a dense one~\cite{hu2020learning}\cite{grzes2009theoretical}; however, it is hard to design the intermediate rewards for all black box models. Hence, we develop a hierarchical curiosity-driven exploration mechanism to P-DQN.

Specifically, we adopt a {\em Random Network Distillation} (RND) module~\cite{burda2018exploration}, i.e., a curiosity-driven approach to generate state curiosity $r^{i,s}_t$ by quantifying the next-state novelty, which will be further combined with the external rewards $r_t$ to generate the modified reward $r'_t=r^{i,s}_t+r_t$. RND transforms the exploring procedure into a supervised learning task by minimizing the mean squared error:
\[r^{i,s}_t=||\hat{f}(s_t,\eta_1)-f(s_t)||^{2},\]
where $f(s_t)$ is the fixed target network and $\hat{f}(s_t,\eta_1)$ is a trainable predictor network learning to distill the target network. 
The loss of trained input state ($s_t$) will decrease as the frequency of visited states increases; therefore, $r^{i,s}_t$ of novel states are expected higher.

Considering that the bonus is obtained after reaching the next state $s_{t+1}$ and will be combined with the environment reward, the agent tends to follow the existing experiences rather than keeps exploring unknown states by taking various actions. 
To encourage the exploration of different actions at each state, we introduce the RND module to reflect the frequency of each action. At state $s_t$, the curiosity of a hybrid action $a_t=(k_t,v_{k_t}^{\params_2}(s_t))$ is given by:
\[r^{i,a}_t=||\hat{g}(s_t,a_t,\eta_2)-g(s_t,a_t)||^{2},\] where $g$ and $\hat{g}$ have the same role as $f$ and $\hat{f}$, respectively. 

Here, we leverage state curiosity $r^{i,s}_t$ to enhance high-level index planning in discrete spaces, and $r^{i,a}_t$ is adopted as low-level action curiosity for enthusiastically searching continuous parameters. Finally, we organically incorporate the two-level curiosity loss into the loss functions defined above in Section~\ref{subsec:policy-opt}:
\[\mathcal{L}'_Q(\params_1,\eta_1)=[Q^{\params_1}(s_{t},k_t,v_{k_t})-y_t]^2+||\hat{f}(s_t,\eta_1)-f(s_t)||^{2},\] \[\mathcal{L}'_{\pi}(\params_2,\eta_2)=-\sum\limits_{k_t\in\features}Q(s_t,k_t,v_{k_t}^{\params_2}(s_t))+||\hat{g}(s_t,a_t,\eta_2)-g(s_t,a_t)||^{2}.\]
It is worth remarking that $a_t$ is a function of $\params_2$. By alternatively updating $\params_2$ and $\eta_2$, the policy network $v_{k_t}^{Q}$ will learn to balance the action's potential value and novelty.

The overview of our {\sc ReLAX} framework is depicted in Figure~\ref{fig:relace}.

\subsection{Global vs. Local Policy}
\label{subsec:global-vs-local}
So far, we have considered one single agent as responsible for generating the CFs for {\em all} the instances of a given dataset, and hereinafter we refer to it as {\sc ReLAX-Global}. 
This allows us to learn a generalized policy that is able to produce counterfactuals, which are not tailored to a specific input sample. 
Algorithm~\ref{alg:relace-global} describes the training of {\sc ReLAX-Global} with the hierarchical curiosity-driven exploration technique discussed in Section~\ref{subsec:Curiosity} above.
\begin{algorithm}[htb!]
\caption{Training {\sc ReLAX-Global} Agent with Curiosity}
\small
\label{alg:relace-global}
\begin{algorithmic}[1]
\State
$\params_1 \gets \textit{initialize the deep }Q\textit{-network } Q^{\params_1}$\\
$\params_2 \gets \textit{initialize the deterministic policy network } v_{k_t}^{\params_2}$ \\
$\eta_1 \gets \textit{initialize the RND state modules } \hat{f}, f$\\ 
$\eta_2 \gets \textit{initialize the RND action modules } \hat{g}, g$ 
\State $\mathcal{M} \gets \textit{initialize the replay buffer}$
\State $i \gets 1$
\While{$i \leq $\text{max\_epochs}}
\State $\inst \sim \dataset$ \Comment{Sample a training instance $\inst$ from $\dataset$}
\State $\inst_0 \gets \inst$
\State $s_0 = (\inst_0, \boldsymbol{f}_0)$ \Comment{Initial state}
\State $t \gets 0$
\For{$t \leq T$}
\Comment{Maximum agent steps for each sample ($T$=50,000)}
\State $v_{k_t} \gets v_{k_t}^{\params_2}(s_t)$ \Comment{Compute the continuous parameter}
\State $a_t \gets (k_t, v_{k_t})$ \Comment{Select the discrete action by $\varepsilon$-greedy}
\State $r^{i,a}_t \gets ||\hat{g}(s_t,a_t,\eta_2)-g(s_t,a_t)||^{2}$\\
\Comment{Generate action curiosity $r^{i,a}_t$}
\State $r_t, s_{t+1} \gets \mathcal{T}(s_t, a_t)$\\ \Comment{The agent gets the reward and observes the next state}
\State $r^{i,s}_t \gets ||\hat{f}(s_t,\eta_1)-f(s_t)||^{2}, \quad r'_t=r^{i,s}_t+r_t$\\
\Comment{Generate state curiosity $r^{i,s}_t$ and modified reward $r'_t$}

\State $p_t \gets \textit{compute the importance $p_t$}$
\State $\mathcal{M} \gets (\{s_t\},\{a_t\},\{r'_t\},\{s_{t+1}\},\{p_t\})$\\
\Comment{Store transition into the replay buffer}
\State $B \sim \mathcal{M}$ \Comment{Randomly sample batch $B$ from $\mathcal{M}$}
\State $\params_1 \gets \params_1 - \gamma_1 \nabla \mathcal{L}'_Q(\params_1,\eta_1), \quad \eta_1 \gets \eta_1 - \gamma_2 \nabla \mathcal{L}'_Q(\params_1,\eta_1) $
\State $\params_2 \gets \params_2 - \gamma_3 \nabla \mathcal{L}'_{\pi}(\params_2,\eta_2), \quad \eta_2 \gets \eta_2 - \gamma_4 \nabla \mathcal{L}'_{\pi}(\params_2,\eta_2) $\\
\Comment{Update the parameters of both networks via SGD}
\State $t \gets t+1$
\EndFor
\State $i \gets i+1$
\EndWhile\\
\Return $\params_1, \params_2$ \Comment{Optimal parameters of both networks}
\end{algorithmic}
\end{algorithm}
However, in some cases, the CF generation process should capture the peculiarities of each individual original instance.
To accomodate such a need, we introduce a variant called {\sc ReLAX-Local}.

{\sc ReLAX-Local} trains a dedicated agent for crafting the optimal CF for a {\em single} target example. It starts by initializing {\sc ReLAX-Local}'s policy with a pretrained {\sc ReLAX-Global}'s policy. Then, using a standard transfer learning approach~\cite{zhuang2021tl}, {\sc ReLAX-Local} is fine-tuned on the target samples.
This step consists of randomly generating synthetic training data points around the target example, using a method similar to~\cite{guidotti2018lore}. 
Specifically, we uniformly sample data points whose $L^2$-norm from the target example is at most $1$.\footnote{Notice that all our input samples are normalized unit vectors.}
%%%%%%%%%%%%%%%%%%%%%%%%%%%%%%%%%%%%%%%%%%%%%%%%%%%%%%%%%%%%%%%%%%%%%%%%
%%%%%%%%%%%%%%%%%%%%%%%%%%%%% EXPERIMENTS %%%%%%%%%%%%%%%%%%%%%%%%%%%%%
\section{Experiments}
\label{sec:experiments}

\subsection{Setup}
\label{subsec:setup}
\noindent {\bf \em Datasets and Tasks.}
We test with five public tabular datasets described in Table~\ref{tab:datasets}, used for classification and regression tasks. 
\begin{table}[ht]
\centering
\footnotesize
\begin{tabular}{|l|c|c|c|}
\hline
\textbf{Dataset}        & \textbf{N. of Instances} & \textbf{N. of Features} & \textbf{Task} \\ \hline
\textit{Breast Cancer}~\cite{breast-ds}  & 699                      & 10 (numerical)          & classification \\ \hline
\textit{Diabetes}~\cite{diabetes-ds}       & 768                      & 8 (numerical)           & classification\\ \hline
\textit{Sonar}~\cite{sonar-ds}          & 208                      & 60 (numerical)          & classification\\ \hline
\textit{Wave}~\cite{wave-ds}           & 5,000                    & 21 (numerical)          & classification \\ \hline
\textit{Boston Housing}~\cite{boston-housing-ds} & 506                      & 14 (mixed)              & regression\\ \hline
\end{tabular}
\caption{\label{tab:datasets}Main characteristics of the five public datasets used.
}
\vspace{-5mm}
\end{table}

\noindent {\bf \em Predictive Models.}
Each dataset is randomly split into 70\% training and 30\% test portions. 
For each task and the associated dataset, we train the suitable set of predictive models chosen amongst the following: Random Forest (RF), Adaptive Boosting ({\sc AdaBoost}), Gradient Boosting ({\sc XGBoost}), Multi-Layer Perceptron (MLP), and Multi-Layer Perceptron for Regression ({\sc MLP-Reg}). 
Both MLP and {\sc MLP-Reg} are fully-connected feed-forward neural networks with two hidden layers; MLP includes also a logistic (i.e., sigmoid) activation function at the last output layer.
Notice that some combinations do not apply, e.g., {\sc MLP-Reg} is only trained on the \textit{Boston Housing} dataset.
We perform 10-fold cross validation on the training set portion of each dataset to fine-tune the hyperparameters of all the trainable models. 
Hence, for each task/dataset pair, we re-train all the applicable models with the best hyperparameters on the whole training set, and we measure their performance on the test set previously hold out. 
To assess the quality of the predictive models, we use accuracy for classification and RMSE for regression.
Eventually, we consider only the best performing model(s) for each task/dataset pair. 
In Table~\ref{tab:Modelinformation}, we summarize the main characteristics of each predictive model used in combination with the benchmarking datasets.
\begin{table}[ht]
\centering
\footnotesize
\begin{tabular}{|l|c|c|}
\hline
\textbf{Dataset {[}Best Model{]}}   & \textbf{Structure} & \textbf{ Acc. ($\blacktriangle$)/RMSE ($\blacklozenge$)} \\ \hline
\textit{Breast Cancer} {[}RF{]} & \{\#trees=100\}  & 0.99 ($\blacktriangle$) \\ \hline
\textit{Diabetes} {[}{\sc AdaBoost}{]} &  \{\#trees=100\} & 0.79 ($\blacktriangle$)\\  \hline
\textit{Wave} {[}{\sc XGBoost}{]}& \{\#trees=100\}  &  0.95 ($\blacktriangle$)\\ \hline\hline
\textit{Breast Cancer} {[}MLP{]}   &  \{\#L1=64, \#L2=128\} &  1.00 ($\blacktriangle$)\\  \hline
\textit{Sonar} {[}MLP{]} & \{\#L1=256, \#L2=256\}  & 0.90 ($\blacktriangle$) \\  \hline
\textit{Wave} {[}MLP{]} & \{\#L1=100, \#L2=200\} & 0.97 ($\blacktriangle$)\\  \hline
\textit{Boston Housing} {[}{\sc MLP-Reg}]& \{\#L1=50, \#L2=128\} & 3.36 ($\blacklozenge$)\\
 \hline
\end{tabular}
\caption{\label{tab:Modelinformation} Model structure and performance for each dataset/task pair.}
\vspace{-4mm}
\end{table}

\vspace{-4mm}
\noindent {\bf \em Counterfactual Generator Baselines.}
We compare {\sc ReLAX} with all the CF generator baselines described in Section~\ref{sec:related-work}. 
{\sc Nearest-CT} is considered the simplest approach. 
Furthermore, we distinguish between {\em model-specific} and {\em model-agnostic} methods. The former include: {\sc FeatTweak} and FOCUS (tree-specific); {\sc DeepFool} and {GRACE} (NN-specific). 
The latter are: LORE, MACE, and DiCE.

\noindent {\bf \em Methodology.} We compare the set of CF generation methods that are suitable for each dataset/target model shown in Table~\ref{tab:Modelinformation}. 
In particular, model-agnostic techniques (including both variants of our {\sc ReLAX}) clearly apply to every setting, whereas model-specific approaches can be tested only when the target model matches (e.g., FOCUS can be used only in combination with tree-based models).
Eventually, we generate a separate CF for each input sample in the test set of every dataset above, using all the CF generators relevant to the setting at hand, i.e., all the model-agnostic methods along with model-specific ones that apply.

\noindent {\bf \em Evaluation Metrics.}
We evaluate the quality of generated CFs according to the following four standard metrics~\cite{verma2020arxiv}: {\em Validity}, {\em Proximity}, {\em Sparsity}, and {\em Generation Time}. 
Validity measures the ratio of CFs that actually meet the prediction goal to the total number of CFs generated:\footnote{Some work consider the complementary metric, which is known as {\em Fidelity} and is equal to (1-{\em Validity}).} the higher the validity the better.
Proximity computes the distance of a CF from the original input sample; in this work, we use $L^1$-norm to measure proximity, and therefore the smaller it is the better.
Sparsity indicates the number of features that must be changed according to a CF, and therefore is equivalent to the $L^0$-norm between a CF and the corresponding original input sample. 
The smaller it is the better, as sparser CFs likely lead to more human-interpretable explanations.
Finally, Generation Time computes the time required to generate CFs, which, clearly, should be as small as possible.
All the metrics above are averaged across all the test input samples.
Moreover, experiments were repeated 5 times and results are expressed as the mean $\pm$ standard deviation.

\subsection{Results}
\label{subsec:results}

\begin{figure*}[ht]
\centering
\includegraphics[width=.8\textwidth]{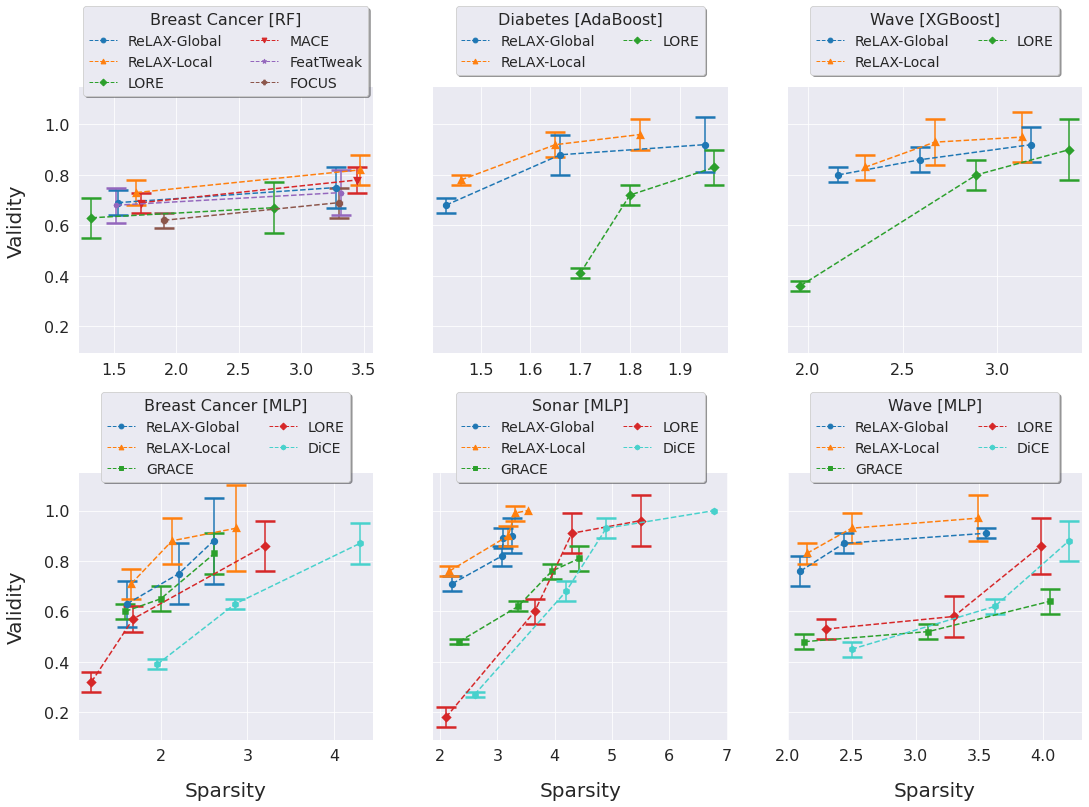}
\caption{{\em Sparsity} vs. {\em Validity} of counterfactuals generated by {\sc ReLAX} and other baselines for classification tasks.}
\label{fig:sparsity-vs-validity-class}
\end{figure*}

\noindent {\bf \em Sparsity-Validity Trade-off.} In Figure~\ref{fig:sparsity-vs-validity-class}, we plot the number of perturbed features (i.e., sparsity) versus the validity of counterfactuals obtained with different CF generation methods, when applied to classification tasks.
More specifically, we fix a threshold on the maximum number of features that each CF generator is allowed to perturb and we show: {(\em i)} the {\em actual} sparsity; and {(\em ii)} the validity of the generated CFs.
The rationale of this analysis is to show which method is able to achieve the best trade-off between two contrasting metrics: sparsity and validity. Intuitively, the larger is the number of perturbed features, the higher is the chance of obtaining a valid counterfactual. 
On the other hand, we should privilege sparser CFs, i.e., CFs that modify as less features as possible, since those are possibly more interpretable and feasible to implement.
Results show that both {\sc ReLAX-Global} and  {\sc ReLAX-Local} achieve the best balance between sparsity and validity. 
That is, our method outperforms all the baselines in terms of validity and, more importantly, it obtains these results even when we set a very restrictive threshold on sparsity, i.e., when few features are modified. 
As expected, though, if we soften such a cap on the number of features allowed to change, other methods like LORE may eventually match the performance of {\sc ReLAX} or even reach higher validity scores. In fact, not controlling for sparsity will make all CF generation methods behave similarly. 
This is what we observe when we test with {\sc DeepFool} or the simplest {\sc Nearest-CT} baseline, which by design tend to generate valid CFs only if a large fraction of the input features get modified. 
Due to this behavior, both {\sc DeepFool} and {\sc Nearest-CT} cannot be visualized in the plots, as they fall outside the range of sparsity values imposed in our experiments.
Furthermore, although MACE is model-agnostic, its applicability to neural network target models is problematic due to its large computational cost~\cite{karimi2020aistats}. 
Besides, the only implementation of MACE remaining available works only in combination with RF target models, and this is why we used it only in the first setting ({\em Breast Cancer} [RF]).

A similar analysis on the sparsity vs. validity trade-off for the {\em Boston Housing} regression task is shown in Table~\ref{tab:sparsity-validity-regr}. In this case, we compare only our two variants of {\sc ReLAX} since none of the CF generation baselines considered is designed to operate in a regression setting. 
As expected, the larger is the tolerance $\delta$ used to determine if the prediction goal of the counterfactual example is met the harder is for {\sc ReLAX} to find a valid counterfactual. 

\begin{table}[ht]
\centering
\footnotesize
\begin{tabular}{c|c|c|}
\cline{2-3}
 & \multicolumn{2}{c|}{\bf \emph{Validity (Sparsity)}}\\
 \hline
 \multicolumn{1}{|c|}{\textbf{Threshold ($\delta$)}} & {\sc ReLAX-Global} & {\sc ReLAX-Local}\\
 \hline
\multicolumn{1}{|c|}{0.20} & $0.81\pm 0.09$ ($\mathbf{3.02}\pm 0.17$) & $\mathbf{0.87}\pm 0.05$ ($3.10\pm 0.18$) \\ \hline
\multicolumn{1}{|c|}{0.40} & $0.74\pm 0.06$ ($\mathbf{3.09}\pm 0.16$) & $\mathbf{0.81}\pm 0.05$ ($3.18\pm 0.16$) \\ \hline
\multicolumn{1}{|c|}{0.60} & $0.70\pm 0.06$ ($\mathbf{3.21}\pm 0.12$) & $\mathbf{0.77}\pm 0.03$ ($3.28\pm 0.09$) \\ \hline
\end{tabular}
\caption{{\em Sparsity} vs. {\em Validity} of counterfactuals generated by {\sc ReLAX} for the {\em Boston Housing} regression task.}
\label{tab:sparsity-validity-regr}
\vspace{-6mm}
\end{table}

Finally, analogous conclusions can be drawn if we compare proximity vs. validity (see Table~\ref{tab:proximity-gentime}): {\sc ReLAX} is able to strike the best balance also between those two conflicting metrics.

\noindent {\bf \em Generation Time.} 
Although validity and sparsity (proximity) are crucial to measure the quality of a CF generation method, efficiency is pivotal as well. 
Therefore, we compare the average generation time for each model-agnostic CF generator, namely LORE, MACE, and our {\sc ReLAX} in its two variants.
We focus on model-agnostic methods because we want this comparison to be as general as possible. %, since model-specific methods do not apply to every setting.
Table~\ref{tab:proximity-gentime} shows that our method takes up to 42\% less time than other model-agnostic baseline to produce valid counterfactuals, which happen to be also closer to the original instances.  
This result is even more remarkable if we consider that the counterfactual generation time of {\sc ReLAX} includes the training of the DRL agent.

\begin{table*}[ht]
\centering
\footnotesize
\begin{tabular}{|c|c|c|c|c|c|}
\hline
\multirow{2}{*}{\textbf{Metric}} & \multirow{2}{*}{\textbf{Dataset} {[}\textbf{Models}{]}} & \multicolumn{4}{c|}{\textbf{CF Generation Methods}}                              \\ \cline{3-6} 
                            &                                        & {\sc ReLAX-Global}    &  {\sc ReLAX-Local}    & LORE   & MACE  \\ \hline
\multirow{5}{*}{{\bf \em Proximity}}    
                             & \textit{Breast Cancer} {[}RF, MLP{]}            & 
                        {[}$\mathbf{4.46},5.92${]}   & {[}$4.49,5.87${]} &
                        {[}$4.63,\mathbf{5.63}${]}   & {[}$4.47$, N/A{]}         \\ \cline{2-6} 
                            & \textit{Diabetes} {[}{\sc AdaBoost}{]}            & {[}$\mathbf{4.41}${]}   & {[}$4.50${]} & {[}$4.76${]}   & {[}N/A{]}      \\ \cline{2-6} 
                            & \textit{Sonar} {[}MLP{]}                        &  {[}$\mathbf{7.32}${]}     & {[}$7.66${]} &
                            {[}$7.36${]}     & {[}N/A{]}     \\ \cline{2-6} 
                            & \textit{Wave} {[}{\sc XGBoost}, MLP{]}            &  {[}$\mathbf{5.93},\mathbf{6.38}${]} & {[}$6.02,6.50${]} &
                            {[}$6.60,6.41${]} & {[}N/A, N/A{]} \\ 
                    \cline{2-6}
                    & \textit{Boston Housing} {[}{\sc MLP-Reg}{]}      &      {[}$\mathbf{5.10}${]}     & {[}$5.36${]} & 
                    {[}N/A{]}     & {[}N/A{]}\\
                            \hline
\hline
\multirow{1}{*}{{\bf \emph{Generation Time} (secs.)}}
                            & \textbf{*}            & {$1500$}   & {$\mathbf{1320}$} & {$2100$}   & {$2280$}   \\ \hline
\end{tabular}
\caption{Comparison of {\em Proximity} and {\em Generation Time} for model-agnostic CF generation methods.}
\label{tab:proximity-gentime}
\vspace{-6mm}
\end{table*}

\subsection{Hyperparameter Tuning}
\label{subsec:ht-tuning}

Our CF generation method is associated with a number of controlling parameters. To understand the impact of these parameters, we analyze the behavior of {\sc ReLAX} on the {\em Sonar} dataset with MLP as the target classifier and the maximal sparsity limited to 5 features.

\begin{figure}[ht]
\centering
\includegraphics[width=\columnwidth]{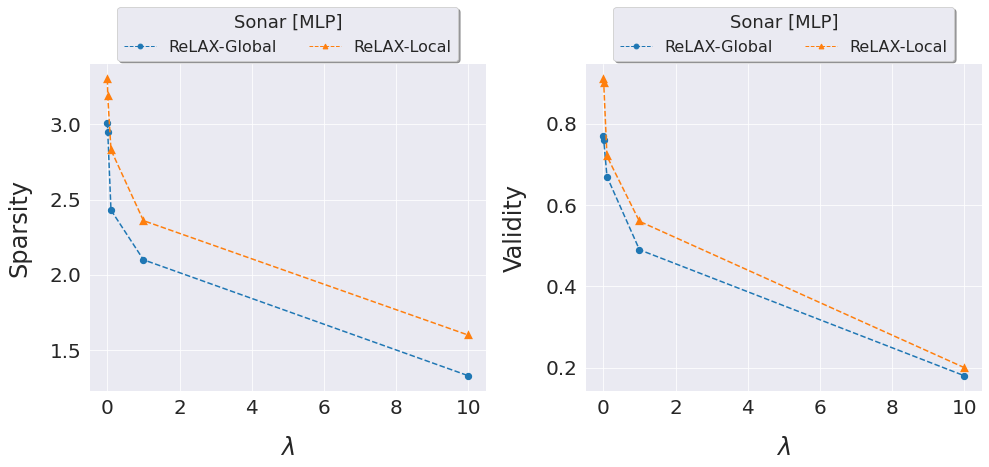}
\caption{The effect of $\lambda$ on \emph{Sparsity} (left) and \emph{Validity} (right).}
\label{fig:ablation-lambda}
\end{figure}

\noindent {\bf \em The scaling factor $\lambda$.}
We first investigate the effect of the scaling factor $\lambda$ on the generated CFs, picking it from ${0.001,0.01,0.1,1,10}$. As shown in Figure~\ref{fig:ablation-lambda}, $\lambda$ controls the balance between sparsity and validity. 
Indeed, larger values of $\lambda$ force the agent to prefer sparser CFs at the expense of lower validity (see Figure~\ref{fig:ablation-lambda} -- left), whereas smaller values of $\lambda$ result in the opposite (see Figure~\ref{fig:ablation-lambda} -- right).

\noindent {\bf \em The target model's architecture.}
We assess the robustness of our CF generation method when applied to different target neural network architectures.
More specifically, Table~\ref{tab:NN architecture} shows the impact of different MLP architectures on the validity and sparsity of counterfactuals generated by {\sc ReLAX} in comparison with two competitors: GRACE (NN-specific) and LORE (model-agnostic).
\begin{table}[ht]
\centering
\footnotesize
\begin{tabular}{lc|c|c|c|c|}
\cline{3-6}
 & & \multicolumn{4}{c|}{\bf \emph{Validity (Sparsity)}}\\
 \hline
 \multicolumn{1}{|l|}{\textbf{MLP size}}& \textbf{Acc.} & {\sc ReLAX-Global} & {\sc ReLAX-Local} & {GRACE} & {LORE}\\
 \hline
\multicolumn{1}{|l|}{[256, 256]} & $0.88$ & $0.76$ ($\mathbf{2.95}$) & $\mathbf{0.90}$ ($3.19$) & $0.62$ ($3.32$) & $0.60$ ($3.65$) \\ \hline
\multicolumn{1}{|l|}{[128, 128]} & $0.85$ & $0.80$ ($\mathbf{2.69}$) &$\mathbf{0.90}$ ($2.88$)  & $0.73$ ($2.85$) & $0.76$ ($3.21$) \\ \hline
\multicolumn{1}{|l|}{[64, 64]} &  $0.79$& $0.90$ ($\mathbf{1.88}$) & $\mathbf{0.95}$ ($1.93$) & $0.86$ ($1.90$) & $0.90$ ($2.52$)\\ \hline
\end{tabular}
\caption{\label{tab:NN architecture} The impact of different MLP architectures on the {\em Validity} ({\em Sparsity}) of counterfactuals.}
\vspace{-4mm}
\end{table}
We may observe that both GRACE and LORE are quite sensitive to the MLP size, i.e., when the target neural network is getting large the validity and sparsity of CFs generated with those two methods deteriorate significantly.
In the case of GRACE, the reason for that detrimental effect is due to the fact that it leverages the gradient of the function approximated by the neural network, which is obviously correlated with the complexity of the MLP structure.
Despite model-agnostic, LORE requires to accurately learn a locally-interpretable surrogate model of the target neural network, which may be hard when this becomes too complex.
Instead, {\sc ReLAX} is more robust toward different MLP structures, and its performance is consistent independently of the MLP size. 
Indeed,  the agent underneath {\sc ReLAX} {\em truly} treats the target model as a black box regardless of its internal complexity.

\noindent {\bf \em The efficiency of pretraining.}
Finally, we show how pretrained {\sc ReLAX-Global} improves the performance of {\sc ReLAX-Local}. 
More specifically, we train {\sc ReLAX-Local} as described in Section~\ref{subsec:global-vs-local}, and compare it with another agent learned from scratch. 
Unsurprisingly, initializing {\sc ReLAX-Local} with pretrained {\sc ReLAX-Global} reduces its CFs generation time, as shown in Table~\ref{tab:Pretrain}.

\begin{table}[ht]
\centering
\footnotesize
\begin{tabular}{|l|c|c|}
\hline
{\bf Setting} & {\bf \emph{Validity}} ({\bf \emph{Sparsity}})& {\bf \emph{Gen. Time} (secs.)}  \\\hline
without pretraining & $0.80$ ($3.27$) &$1132$  \\\hline
with pretraining & $\mathbf{0.90}$ ($\mathbf{3.19}$) & $\mathbf{500}$\\\hline

\end{tabular}
\caption{\label{tab:Pretrain} 
Performance of {\sc ReLAX-Local} with/without pretrained {\sc ReLAX-Global}.}
\vspace{-4mm}
\end{table}
%%%%%%%%%%%%%%%%%%%%%%%%%%%%%%%%%%%%%%%%%%%%%%%%%%%%%%%%%%%%%%%%%%%%%%%%
%%%%%%%%%%%%%%%%%%%%%%%%%%%%% CASE STUDY %%%%%%%%%%%%%%%%%%%%%%%%%%%%%
\vspace{-4mm}
\section{Case Study: COVID-19 Mortality}
\label{sec:casestudy}
The goal of this case study is to demonstrate that counterfactual explanations provided by {\sc ReLAX} may help countries that suffer from high COVID-19 mortality taking effective actions, economically and medically, to reduce such risk. 

To achieve that, we first need to learn a binary classifier that predicts whether the country's death risk due to COVID-19 is high or not, given the country-level features.
Inspired by previous work addressing a similar task~\cite{bird2020country}, we gather 17 country-level demographic features from 156 countries across 7 continents~\cite{Worldometers,CIA,WHOwf,WHOobe}. 
Furthermore, we label each instance with a binary class denoting the COVID-19 risk of mortality (``high'' vs. ``normal''), as in~\cite{bird2020country}.
This dataset is made publicly available.\footnote{\url{\repourl}}

\begin{figure}[ht]
\centering
\includegraphics[width=.9\columnwidth]{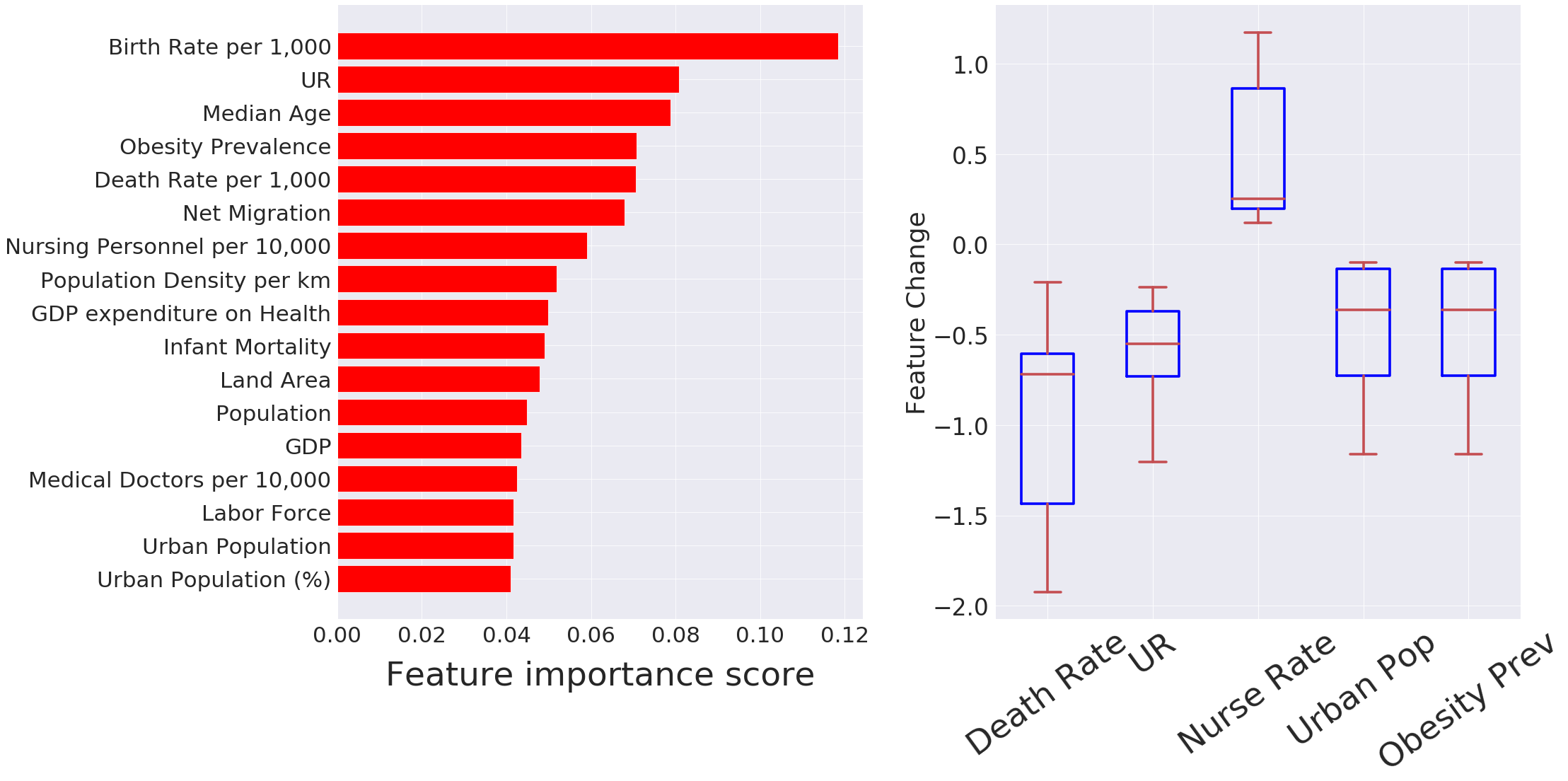}
\caption{Left: Ranking of feature importance. Right: Suggested feature changes for lowering high mortality risk.}
\label{fig:covid-19-features}
\vspace{-4mm}
\end{figure}

We randomly split our collected dataset into two parts, i.e., $70\%$ used for training and $30\%$ for test.
Therefore, we train the following models to learn the binary classifier for predicting COVID-19 mortality risk: SVM, RF, {\sc AdaBoost}, and {\sc XGBoost}.
After running 10-fold cross validation, the best-performing model turns out to be {\sc XGBoost} with 500 trees, which achieves $85\%$ accuracy on the test set.
We sort the features according to the ranking induced by the {\sc XGBoost} model, as shown in Figure~\ref{fig:covid-19-features} (left). 
Then, we generate with our {\sc ReLAX} algorithm the CFs for the 16 high-risk countries in the test set.
It is worth remarking that, although the target binary classifier is learned considering {\em all} the features from the whole training set, we force {\sc ReLAX} to tweak only the subset of {\em actionable} features so that effective suggestions can be found by observing the generated CFs. Besides, in order to avoid bizarre recommendations, we constrain the change of (actionable) features toward a ``plausible'' direction (e.g., it would not make any sense to suggest increasing the unemployment rate of a country). 

The {\sc ReLAX} agent tweaks 1.67 features on average, and the average proximity (i.e., $L^1$-norm) between the original sample and the corresponding counterfactual example is 1.18. Figure~\ref{fig:covid-19-features} (right) shows the direction of feature changes as suggested by the CFs. 

We observe that the CFs suggest five main changes: \emph{(i)} Decreasing the death rate;\footnote{Albeit it may sound odd, reducing the death rate can subsume a broader suggestion for improving the life quality of a country.} \emph{(ii)} Decreasing the unemployment rate; \emph{(iii)} Increasing the nurse rate per 10,000 people; \emph{(iv)} Decreasing the urban population rate; and \emph{(v)} Decreasing the obesity prevalence.
Not only those recommendations look sensible, as much as straightforward or even obvious, but their accuracy is also confirmed by the fact that many countries have indeed adopted similar strategies to counter the impact of COVID-19.
For example, US approved the visa for more than 5,000 international nurses to strengthen the health workforce.\footnote{\url{https://www.npr.org/sections/health-shots/2022/01/06/1069369625/short-staffed-and-covid-battered-u-s-hospitals-are-hiring-more-foreign-nurses}} 
Moreover, according to the investigation made by the US Center for Disease Control and Prevention,\footnote{\url{https://www.cdc.gov/obesity/data/obesity-and-covid-19.html}} obese patients with COVID-19 aged 18 years and younger are associated with a 3.07 times higher risk of hospitalization and a 1.42 times higher risk of severe illness. Thus, reducing the obesity prevalence can indeed lower mortality risk.
Finally, reducing the unemployment rate and the urban population rate may allow a broader range of people to enhance their social community awareness, take consciousness of the risks of the pandemic, and thus adopt a safe lifestyle.

%%%%%%%%%%%%%%%%%%%%%%%%%%%%%%%%%%%%%%%%%%%%%%%%%%%%%%%%%%%%%%%%%%%%%%%%
%%%%%%%%%%%%%%%%%%%%%%%%%%%%% CONCLUSION %%%%%%%%%%%%%%%%%%%%%%%%%%%%%
\vspace{-2mm}
% CONCLUSION
\section{Conclusion and Future Work}
\label{sec:conclusion}
In this work, we presented {\sc ReLAX}, the first method for generating model-agnostic counterfactual examples based on deep reinforcement learning with hierarchical curiosity-driven exploration. 
We implemented two variants of it: {\sc ReLAX-Global} and {\sc ReLAX-Local}. 
The former learns a generalized agent's policy from a whole set of training instances, whereas the latter trains a dedicated agent's policy for crafting the optimal counterfactual of a single target example via transfer learning from a pretrained {\sc ReLAX-Global}.
Extensive experiments run on five public tabular datasets demonstrated that {\sc ReLAX} significantly outperforms all the considered CF generation baselines in every standard quality metric. 
Our method is scalable with respect to the number of features and instances, and can explain {\em any}  black-box model, regardless of its internal complexity and prediction task (i.e., classification or regression).
Finally, we show that CFs generated by {\sc ReLAX} are useful in practice, as they can be used to suggest actions that a country should take to reduce the risk of COVID-19 mortality.

Several research directions are worth exploring in future work. For example: extend {\sc ReLAX} to generate explanations for non-tabular input data like images, and compare our method against non-counterfactual explanation approaches like SHAP~\cite{lundberg2017shap} or LIME~\cite{ribeiro2016lime}.
%%%%%%%%%%%%%%%%%%%%%%%%%%%%%%%%%%%%%%%%%%%%%%%%%%%%%%%%%%%%%%%%%%%%%%%%

%%%%%%%%%%%%%%%%%%%%%%%%%%%%% ACKNOWLEDGMENTS %%%%%%%%%%%%%%%%%%%%%%%%%%%%%
\vspace{-2mm}
\begin{acks}
This research was supported by the Italian Ministry of Education, University and Research (MIUR) under the grant "Dipartimenti di eccellenza 2018-2022" of the Department of Computer Science and the Department of Computer Engineering at Sapienza University of Rome. 
\end{acks}
%%%%%%%%%%%%%%%%%%%%%%%%%%%%%%%%%%%%%%%%%%%%%%%%%%%%%%%%%%%%%%%%%%%%%%%%

\balance

\end{document}